\documentclass[10pt,twocolumn,letterpaper]{article}

\usepackage[pagenumbers]{wacv} % To force page numbers, e.g. for an arXiv version

\usepackage{graphicx}
\usepackage{amsmath}
\usepackage{amssymb}
\usepackage{booktabs}
\usepackage{array}
\usepackage{booktabs}

\usepackage[pagebackref,breaklinks,colorlinks]{hyperref}

\usepackage[capitalize]{cleveref}
\Crefname{fig}{Fig.}{Figures}
\Crefname{eqn}{Eqn.}{Eqs.}
\Crefname{section}{Sect.}{Sects.}
\Crefname{table}{Table}{Tables}

\def\wacvPaperID{1383} % *** Enter the WACV Paper ID here
\def\confName{WACV}
\def\confYear{2025}

\begin{document}

% create commands for text superscript and subscript
\newcommand{\ts}{\textsuperscript}
\newcommand{\tsb}{\textsubscript}

% creates thick column vertical line
\newcolumntype{?}{!{\vrule width 2pt}} 

\title{Effects of Common Regularization Techniques on Open-Set Recognition}

% \and
\author{
Zachary Rabin$^1$, Jim Davis$^1$, Benjamin Lewis$^{2}$, Matthew Scherreik$^{2}$\\
$^1$Ohio State University, $^2$Air Force Research Laboratory
}
\maketitle

%%%%%%%%% ABSTRACT
\begin{abstract}
    In recent years there has been increasing interest in the field of Open-Set Recognition, which allows a classification model to identify inputs as ``unknown'' when it encounters an object or class not in the training set.
    This ability to flag unknown inputs is of vital importance to many real world classification applications.
    As almost all modern training methods for neural networks use extensive amounts of regularization for generalization, it is therefore important to examine how regularization techniques impact the ability of a model to perform Open-Set Recognition.
    In this work, we examine the relationship between common regularization techniques and Open-Set Recognition performance.
    Our experiments are agnostic to the specific open-set detection algorithm and examine the effects across a wide range of datasets.
    We show empirically that regularization methods can provide significant improvements to Open-Set Recognition performance, and we provide new insights into the relationship between accuracy and Open-Set performance.
\end{abstract}

%%%%%%%%% BODY TEXT
\section{Introduction}
\label{sec:intro}

There is a trend towards larger and deeper neural networks to achieve high levels of classification accuracy, but oftentimes they require forms of regularization during training.
In many cases, the goal of added regularization is to prevent overfitting in order to improve generalization and overall test accuracy.
Alternatively, regularization can be used to induce other behaviors in a model such as better calibration, robustness to adversarial attacks, etc.

After training with added regularization, models are put into real world use.
When predicting a label for an input, a neural network is forced to choose from a set of known labels as its output.
These known labels are the set of classes exposed to the network during training.
This paradigm works so long as we can assume that all inputs to the network belong to one of the classes in that known set.
However, in many real world applications, this assumption can often be violated.

One could attempt to alleviate this issue by rejecting unconfident predictions.
However, it is possible that an unknown object exists within the same classification space as a known object or in a space that produces erroneously high confidence.
This motivates the need for Open-Set Recognition.

In Open-Set Recognition (OSR), the goal is to filter out unknown or novel inputs while  retaining high accuracy on examples of the known labels.
As previously mentioned, regularization schemes are often used to improve generalization. 
However, it is conceivable that such regularization schemes could potentially cause issues for OSR by generalizing in such a way as to confidently classify examples outside of the training domain.

Therefore, it is important to examine the effect, if any, of regularizers to OSR.
We examine classic L2 regularization (weight decay), label smoothing \cite{LabelSmoothing}, Mixup \cite{Mixup} (which was shown to have an approximate form of regularization \cite{MixupAsReg}), and CutMix \cite{CutMix} (also shown to be a form of regularziation \cite{OnMixup, MixupAsReg, CutMixReg}). 
In this work we focus on the distances between penultimate feature vectors of known and unknown examples to study the effects of regularization on OSR.

%------------------------------------------------------------------------
\section{Related Work}
\label{sec:related}

The concept of OSR was first formalized in \cite{towardOpenset}.
This paper introduced the idea that the label spaces of training and testing data could potentially be different.
Training and testing data take the form $\mathit{D\tsb{train}} = \left \{ \left ( x_{i}, y_{i} \right ) \right \}_{i=1}^{N}$ and $\mathit{D\tsb{test}} = \left \{ \left ( x_{i}, y_{i} \right ) \right \}_{i=1}^{M}$, respectively, where $x_{i}$ denotes an example and $y_{i}$ denotes its corresponding ground truth label.
We let $Y\tsb{train}$ and $Y\tsb{test}$ be the set of possible classes from $D\tsb{train}$ and $D\tsb{test}$.
For standard training and evaluation, it holds that $Y\tsb{train} = Y\tsb{test}$.
This scenario is called ``closed-set'' recognition.
Conversely, in the open-set scenario, $Y\tsb{train} \subset Y\tsb{test}$.
If a testing tuple $\left ( x_{i}, y_{i} \right )$ has a ground truth label that satisfies $y_i \in Y\tsb{train}$ and $ y_i \in Y\tsb{test} $ it is a closed-set example.
If the testing tuple has a ground truth label $y_i \in Y\tsb{test}$ but $y_i \notin Y\tsb{train}$ it is an open-set example.
% In general, we can now refer to the closed-set as the group of all closed-set examples.
% Similarly, the open-set is the group of all open-set examples.

There are many approaches to tackling the open-set problem.
A simple baseline approach is described in \cite{baselineOSR} where they show that open-set examples tend to have lower maximum softmax values than closed-set examples, allowing for basic thresholding  of the maximum softmax value.
Thresholding using the maximum softmax value has become a common practice for many OSR algorithms.
For example, the ODIN framework described in \cite{ODIN} uses temperature scaling and adversarial noise to utilize the maximum softmax scores for OSR.

Another approach in OSR is OpenMax \cite{OpenMax}.
Instead of using the maximum value from the softmax output, OpenMax models the entire softmax distribution to determine open-set from closed-set examples.
They argue that while it is possible for open and closed-set examples to have similar maximum softmax values, the distribution of \emph{all} softmax values will be meaningfully different.
Therefore they analyze the distances to the average softmax vector per class as their detection algorithm.

In \cite{ClosedSet} it was shown that given sufficient increases in closed-set accuracy, thresholding using softmax values can outperform other advanced algorithms.
They used deeper models as well as more extended training and showed that these strongly correlate to better OSR.
Additionally, it was highlighted that the increases in accuracy brought about by deeper models and enhanced training procedures had a strict linear trend with the OSR capabilities of the model.

Another approach is the GODIN algorithm \cite{GODIN}, which takes advantage of conditional probabilities to estimate the chances of an example being open-set.
This is done using two additional linear layers at the end of the network that are encouraged to represent the probabilities of examples belonging to the closed-set or open-set.
Instead of doing detection at the softmax layer, the GODIN algorithm performs detection at the end of the additional layers.
Alternatively, Reciprocal Point Learning (RPL) \cite{RPL} and Adversarial Reciprocal Point Learning (ARPL) \cite{ARPL} perform detection at the penultimate feature layer. 

These previous OSR works mainly focus on techniques and training strategies that maximize the OSR capabilities of models.
We instead examine a particular component of training commonly used in neural networks, regularization, and relate its use to OSR.
While other work highlight specific trends between accuracy and the performance of OSR algorithms \cite{ClosedSet}, we will show that regularization does not follow this trend.
Additionally, we will show how the feature spaces of models are altered when using regularization techniques to allow for better open-set filtering.

%------------------------------------------------------------------------
\section{Regularization}
\label{sec:regularization}

Regularization methods for classification in machine learning commonly take the form of an additive term to a standard classification loss \cite{Reg}. Here we present our definition of regularization.

\noindent \textbf{Definition 1: Regularization.
} \emph{Enforcement of a constraint on the learning of a model.
The constraint comes from prior knowledge of favorable behaviors or states within a network.
To achieve these constraints there exists a loss function of the form $\hat{L} = L + \lambda R$ where $L$ is the original loss function for the task, $R$ is a regularization term enforcing the constraints, and $\lambda$ is a weighting factor on the regularization.}

In this section, we will present four common regularization strategies to be examined and discuss how each fits our definition of regularization and their potential implications on OSR. 

\subsection{L2 Regularization}
\label{ssec:l2}
L2 regularization \cite{WeightDecay}, is a common technique that is used when training many modern neural network classifiers.
This regularization scheme adds a penalty term to the loss which is composed of the sum of squared weights in the network

\begin{equation}
    L_2 = L + \frac{\lambda}{2N} \sum_i^N w_i^2
    \label{L2.eq}
\end{equation}

\noindent The penalty term enforces that the weights in the network remain small.
When used with stochastic gradient descent the effect of L2 regularization is equivalent to ``weight decay''

\begin{equation}
    w_i = \left (1-\eta \frac{\lambda}{N} \right )w_i - \eta \frac{\partial L}{\partial w_i}
    \label{weightdecay.eq}
\end{equation}

L2 regularization, and by extension weight decay, is typically only applied to the weights of the neural network and excludes all biases.

\subsection{Label Smoothing}
\label{ssec:labelsmooth}
Label smoothing \cite{LabelSmoothing} works by augmenting the output target vectors with noise.
This added noise is also used to help calibrate the resulting model.
The new target vector $q^{\prime}$ is formed as a convex combination of the original one-hot target vector $q$ and a noise vector $u$ that represents a smoothing distribution to be applied

\begin{equation}
    q^{\prime}= (1-\alpha) q + \alpha u
    \label{labelsmoothing.eq}
\end{equation}

\noindent In practice, $u$ is chosen to be a uniform vector with elements $\frac{1}{K}$ where $K$ is the number of classes in the closed set.
Another way to look at this operation is to think of how it affects the cross entropy loss $H$ for the target $q^\prime$ and output vector $p$ 

\begin{equation}
    H(q^\prime,p) = (1-\alpha)H(q,p) + \alpha[D_{KL}(u||p) + H(u)] 
    \label{labelsmoothingloss.eq}
\end{equation}

\noindent This equivalence was shown in \cite{whenLabelSmooth}, and the operation is composed of the standard cross entropy loss along with the KL divergence \cite{KLDivergence} between the smoothing distribution and the output.
The KL divergence term is used to enforce that output vectors must not be close to one-hot vectors.
As the noise vector $u$ is fixed, $H(u)$ is a constant and can be removed from the loss without affecting the gradients. 
Rearranging the remaining terms yields a label smoothing loss

\begin{equation}
    L_{ls} = H(q,p) + \frac{\alpha}{1-\alpha}D_{KL}(u||p)
    \label{lsarrange.eqss}
\end{equation}

\noindent that matches the form shown in our definition of regularization. 

\subsection{Mixup}
\label{ssec:mixup}
For image classification, Mixup \cite{Mixup} is a strategy that works as a random convex blend of input images and also their corresponding output one hot label vectors. 
An example for the images in Figs.~\hyperref[fig:pics]{1(a)} and \hyperref[fig:pics]{1(b)} can be seen in Fig.~\hyperref[fig:pics]{1(c)}.
When applied to neural networks, Mixup can be used to improve both classification accuracy and generalizability \cite{Mixup}.
Multiple papers have shown that the Mixup loss can be closely approximated by the cross entropy loss with regularizing terms of the gradients \cite{OnMixup, MixupAsReg}.

\subsection{CutMix}
\label{ssec:cutmix}
\begin{figure}
\scriptsize
\begin{center}
\begin{tabular}{cc}
    \includegraphics[width=0.45\linewidth]{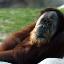} & \includegraphics[width=0.45\linewidth]{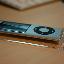} \\
    (a) Image A & (b) Image B \\
    \includegraphics[width=0.45\linewidth]{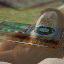} & \includegraphics[width=0.45\linewidth]{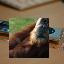} \\
    (c) Mixup Image & (d) CutMix Image
\end{tabular}
\end{center}
\caption{Example of Mixup and CutMix on Tiny ImageNet data.} \label{fig:pics}
\end{figure}

CutMix \cite{CutMix} is a popular extension of Mixup that cuts a rectangular region from one image and pastes it onto another image. 
An example for the previous images can be found in Fig.~\hyperref[fig:pics]{1(d)}.
Recent works have shown that the CutMix strategy also has relations to regularizing the first and second derivatives of the model with respect to its inputs \cite{CutMixReg}.

\section{Experimental Setup}
\label{sec:setup}

One could ask whether these types of regularization constraints could hurt a model's ability to handle open-set examples during inference. 
Stated differently, can regularization generalize the model so much that it begins to accept open-set examples?
We compare the four regularization methods in \cref{sec:regularization} to a baseline model without regularization across multiple architectures and datasets in order to evaluate the effect of these techniques.

We let the term open-set performance (OSP) be the evaluation of a model's ability to distinguish and filter open-set examples from closed-set examples. 
We also let the term closed-set performance (CSP) refer to the model's accuracy on the closed-set testing data.

To perform open-set filtering, some algorithms use the output logits or softmax values, while other algorithms add extra detection layers, use auxiliary networks, or employ penultimate feature representations.
For our evaluation of OSP we choose to examine the penultimate layer of different ResNet \cite{ResNet} architectures, which is the Global Average Pooling (GAP) layer.
If features are more separable at the GAP layer, we would also expect them to be more separable downstream, regardless of the specific loss function.
One could use earlier layers in the network, however deep neural networks typically do not become separable until the last few layers of the network, regardless of network depth \cite{nucleation}.

\subsection{Metrics}
\label{ssec:metrics}
We use two metrics for our OSP evaluation based on the penultimate features of the network.
To calculate our metrics we first compute the GAP feature vectors for all of the closed-set training data (excluding validation).
This will serve to model the closed-set data.
With this, we compute the mean feature vector per class in the training set.
We denote this vector as the class prototype vector.
Secondly, we compute the feature vectors from the test set, which contains both open-set and closed-set data.

A good classifier with high accuracy will have a feature space such that the closed-set examples lie close to their target class's prototype vector and far from the other prototypes.
Additionally, the same classifier with good OSP will have a feature space where open-set examples lie far from all of the class prototypes.
Therefore, scoring vectors on their distance to each prototype provides insight on the ability of the model to separate closed and open-set examples in the feature space.

\subsubsection{Area of Overlap}
\label{sssec:overlap}
Our first open-set metric is a per-class score averaged across the closed-set classes.
For each closed-set class, we have the class prototype and the entire population of testing data feature vectors from that class.
In addition, we have the set of open-set feature vectors from the open-set dataset.

\begin{figure}
\scriptsize
    \centering
    \begin{tabular}{c}
        \includegraphics[width=0.8\columnwidth]{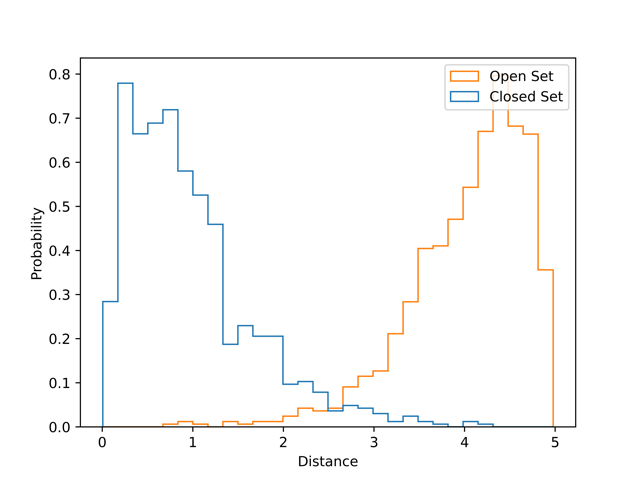} \\
        (a) Desirable low histogram overlap.\\
        \includegraphics[width=0.8\columnwidth]{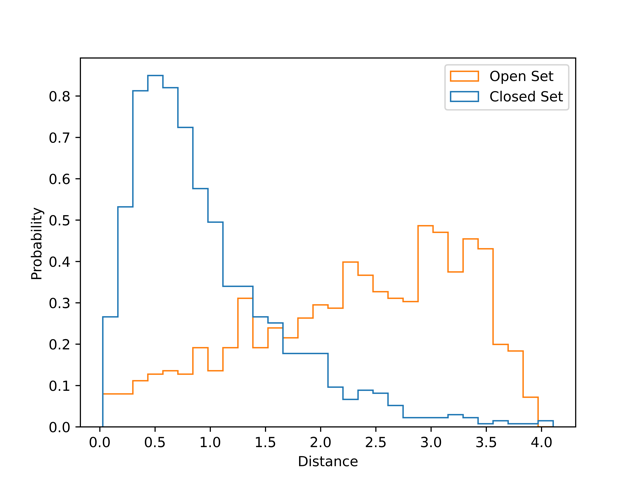} \\
        (b) Undesirable high histogram overlap.
    \end{tabular}
    \caption{Example of desirable and undesirable histogram overlap.}
    \label{fig:histogram}
\end{figure}

For each class we create two histograms of distances to the target class's prototype.
We use the Freedman-Diaconis rule \cite{histogram} to determine the width of the histogram bins.
One histogram is based on the distances of closed-set vectors to the class prototype, while the other uses the distances of open-set vectors to the class prototype.
Both histograms utilize the same bin widths and locations, and are normalized to have an area under the curve of 1. 
Finally, we compute the area of overlap between the two histograms.
A lower area corresponds to a lower probability of finding an open-set and a closed-set vector the same distance away from a class prototype. 
This in turn means that there is better separation between the closed and open-set vectors. 
Higher areas of overlap are therefore undesirable as they correspond to worse separation of open-set vectors. 
Examples of low and high overlaps can be seen in Fig.~\ref{fig:histogram}.
Finally, we take the average of the area of overlap values across all closed-set classes as our final score.

\subsubsection{AUROC}
\label{sssec:auroc}
To calculate the AUROC score we first calculate the Euclidean distance between closed-set testing feature vectors and all of the class prototypes.
We also calculate the distance between the open-set vectors and all of the closed-set class prototypes.
For each example, either closed or open-set, we assign a score that is equal to the minimum distance between that example and any of the class prototypes.
We compare the generated scores to a series of thresholds to identify closed-set and open-set examples.
Examples below the threshold are considered closed-set, while those above are deemed open-set.

For each threshold tested we calculate the true positive rate and the false positive rate of detecting an open-set example.
Plotting these values such that the horizontal axis is the false positive rate and the vertical axis is the true positive rate, we generate a Receiver Operating Characteristic (ROC) curve \cite{ROC}.
Finally, we compute the area under the ROC curve, or AUROC.
A perfect detector will have an AUROC value of 1, while a random detector will have an AUROC of 0.5.

\subsection{Datasets}
\label{ssec:datasets}

In the domain of image classification, an open-set image could be any image that has a label $y$ that is not within $Y\tsb{train}$ (as stated in \cref{sec:related}).
However, in practice, it is impossible to exhaustively sample the space of all images to determine that an algorithm can handle \emph{all} open-set data.
Therefore, when evaluating OSR we choose a sample dataset not used in training to act as the open-set.
We follow \cite{OpenMax, GODIN, ODIN, openmatch, ClosedSet} to create closed and open-set datasets by combining and splitting existing datasets.
We utilize CIFAR10 \cite{CIFAR}, CIFAR100 \cite{CIFAR}, Tiny ImageNet \cite{TinyImageNet}, and Fine Grained Visual Classification Aircraft (FGVCA) \cite{fgvca} to create our datasets.

CIFAR10 and CIFAR100 are benchmark datasets each consisting of 60K, 32x32 color images. 
CIFAR10 has 10 classes, each with 5K training examples and 1K testing examples.
CIFAR100 has 100 classes, each with 500 training examples and 100 testing examples. 
Classes in the datasets are a mix of natural objects (animals, plants, etc.) and man-made objects (airplanes, computers, etc.). 

Tiny ImageNet contains 100K, 64x64 color images, evenly split across 200 classes. 
Each class has 500 training images and 50 validation images. 
Images in this dataset are downsized images from the ImageNet dataset \cite{ImageNet}, which contains man-made and natural objects sourced from the Internet.
As there is no publicly available labeled test set, we use the validation set for the testing set. 

FGVCA contains 10K images of 100 aircraft classes. 
The images vary in size and are evenly split between the training, testing, and validation sets.
We resize the smaller edge of each image to 224 pixels, and then take a center crop of 224x224 so that all images have the same size. 
The FGVCA dataset has a label hierarchy with varying degrees of finer or coarser labels. 
For our experiments we use the family level hierarchy which contains 70 unbalanced classes.

From these benchmark datasets we create 5 open-set datasets.
When creating our datasets we ensure that there is no overlap of classes between the closed and open-set parts.
The created datasets along with their closed-set and open-set parts are provided in \cref{tab:datasets}. 

\begin{table}
    \centering
    \scriptsize 
    \resizebox{\columnwidth}{!}{
    \begin{tabular}{|c?c|c|}
        \hline 
        Dataset & Closed-Set & Open-Set \\
        \hline
        \hline
        CI 6-4 & 6 CIFAR10 classes & 4 CIFAR10 classes \\
        \hline
        CI~10+ & CIFAR10 & CIFAR100 \\
        \hline
        CI~100+ & CIFAR100 & CIFAR10 \\
        \hline
        TIN~100-100 & 100 Tiny ImageNet classes & 100 Tiny ImageNet Classes \\
        \hline
        FGVCA~35-35 & 35 FGVCA classes & 35 FGVCA classes \\
        \hline
    \end{tabular}
    }
    \caption{Datasets and their corresponding closed-set and open-set parts.}
    \label{tab:datasets}
\end{table}

\subsection{Training Process}
Unless specified otherwise, all models were trained using the same procedures listed in this section.
All code was implemented in the PyTorch framework and models were trained using either a Tesla T4 GPU or a V100 GPU.

For datasets CI~6-4 and CI~10+, a ResNet18 was used as the base model.
We used a ResNet34 for the CI~100+ and TIN~100-100 datasets.
Finally, a ResNet50 was used for FGVCA~35-35.
Models were trained for 500 epochs using the Stochastic Gradient Descent (SGD) optimizer with an initial learning rate of 0.1, a cosine half-period learning rate scheduler, and a momentum of 0.9.
When using a ResNet50, we decreased the initial learning rate to 0.01 as it had better validation accuracy.
We employed a batch size of 128 for all datasets except FGVCA~35-35, which instead used a smaller batch size of 8 due to the larger network and image sizes.

Our input augmentation scheme during training consisted of random horizontal flipping, mean padding and random cropping, and color jitter followed by standard data normalization.
We purposely used minimal data augmentation to emphasize the effects of the regularizers examined in this work.

When L2 regularization (weight decay) was used, the loss function was the average cross entropy for each batch.
Additionally, the hyperparameter, $\lambda$ was chosen such that its value times the number of parameters in the network $N$ was a constant value.
We kept $\lambda N$ as a constant so that as $N$ increased with a larger network, $\lambda$ appropriately decreased.
After testing $\lambda$ values in the range [1e-6, 1e-3] for the best validation accuracy, we chose the corresponding constant to be 1100.
When label smoothing was used we followed \cite{LabelSmoothing} and employed an $\alpha$ parameter value of 0.1.
Finally, the Mixup and CutMix blending parameter was chosen randomly from a uniform distribution for each batch.

%------------------------------------------------------------------------
\section{Results}
\label{sec:results}

In this section, we present the results of our proposed experiments. 
In the tables we abbreviate the baseline approach with no regularization as ``Base''.
For approaches with regularization, we abbreviate L2 regularization as ``L2'', label smoothing as ``LS'', Mixup as ``MU'', and CutMix as ``CM'' . 
Unless otherwise specified, all results are computed from the average of 3 randomly initialized training runs.
Additionally, for split datasets (CI 6-4, TIN 100-100, and FGVCA 35-35) we randomly select classes to be part of the open or closed-sets for each training run.

\begin{table}
    \centering
    \scriptsize 
    \begin{tabular}{|c?c?c|c|c|c|}
        \hline
        Dataset & Base & L2 & LS & MU & CM\\
        \hline\hline
             CI 6-4 &  93.31 &  94.61 &  93.13 &  95.23 &  95.03 \\
             \hline
             CI~10+ &  92.97 &  94.76 &  93.16 &  95.30 &  95.27 \\
             \hline
            CI~100+ &  70.01 &  73.75 &  71.06 &  74.81 &  75.11 \\
            \hline
 TIN~100-100 &  62.74 &  65.50 &  62.06 &  66.84 &  67.50 \\
 \hline
        FGVCA~35-35 &  79.37 &  80.65 &  79.13 &  80.76 &  76.41 \\
        \hline
    \end{tabular}

    \caption{Average closed-set accuracy of models on the closed-set test set across three runs.}
    \label{tab:accuracy}
\end{table}

\begin{table}
    \centering
    \scriptsize 
    \begin{tabular}{|c?c?c|c|c|c|}
        \hline
           Dataset &  Base &    L2 &    LS &    MU &    CM \\
                \hline\hline
            CI 6-4      &  0.30 &  0.20 &  0.22 &   0.21 &    0.18 \\ \hline
            CI 10+      &  0.27 &  0.14 &  0.15 &   0.16 &    0.12 \\ \hline
            CI 100+     &  0.36 &  0.15 &  0.20 &   0.31 &    0.32 \\ \hline
            TIN 100-100 &  0.35 &  0.18 &  0.22 &   0.35 &    0.29 \\ \hline
            FGVCA 35-35 &  0.23 &  0.21 &  0.29 &   0.21 &    0.25 \\ \hline
    \end{tabular}
    \caption{Average value of histogram overlap where a \underline{lower} value is considered better.}
    \label{tab:overlap}
\end{table}

\begin{table}
    \centering
    \scriptsize 
    \begin{tabular}{|c?c?c|c|c|c|}
        \hline
            Dataset &  Base &    L2 &    LS &    MU &    CM \\
                \hline\hline
            CI 6-4      &  0.68 &  0.74 &  0.73 &   0.78 &    0.79 \\ \hline
            CI 10+      &  0.65 &  0.88 &  0.80 &   0.83 &    0.89 \\ \hline
            CI 100+     &  0.55 &  0.73 &  0.74 &   0.69 &    0.73 \\ \hline
            TIN 100-100 &  0.53 &  0.69 &  0.68 &   0.63 &    0.69 \\ \hline
            FGVCA 35-35 &  0.65 &  0.66 &  0.57 &   0.68 &    0.61 \\ \hline
    \end{tabular}
    \caption{Average AUROC values for the basic filtering process where a \underline{higher} value is considered better.}
    \label{tab:auroc}
\end{table}

\subsection{Closed-Set Performance}
\label{ssec:csp}

\cref{tab:accuracy} records the accuracy on the closed-set test set for each model and dataset.
Predictably, we see that added regularization helped to increase the overall CSP.
L2 regularization added significant improvements over the baseline model.
Some datasets showed improvement with label smoothing, while on other datasets label smoothing was only able to roughly match baseline accuracy.
Mixup and CutMix well outperformed the baseline model, except for CutMix on FGVCA~35-35.
We suspect that when using CutMix on FGVCA~35-35 the paste operation could potentially obscure highly relevant information about the underyling class label, for example occluding a plane engine.
Overall, the regularization schemes tested generally helped increase accuracy values across datasets, as expected.

\subsection{Open-Set Performance}
\label{ssec:osp}

\cref{tab:overlap} shows the histogram area of overlap values for the different datasets and models.
A lower value is more favorable as it shows less confusion between the open and closed-set.
Across the different datasets, we observed that the regularization methods examined \underline{improved} the OSP of the algorithm.

\begin{figure}[t]
    \centering
    \includegraphics[width=0.8\columnwidth]{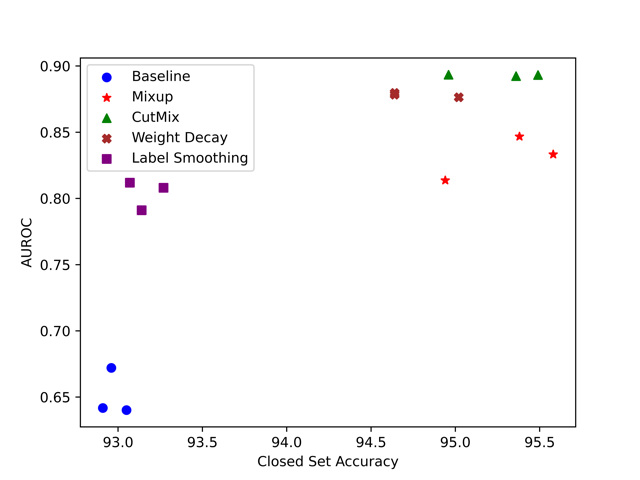}
    \caption{Closed-set accuracy vs AUROC for the CI~10+ dataset across three runs.}
    \label{fig:accVaurocCif}
\end{figure}

\begin{figure}
    \centering
    \includegraphics[width=0.8\columnwidth]{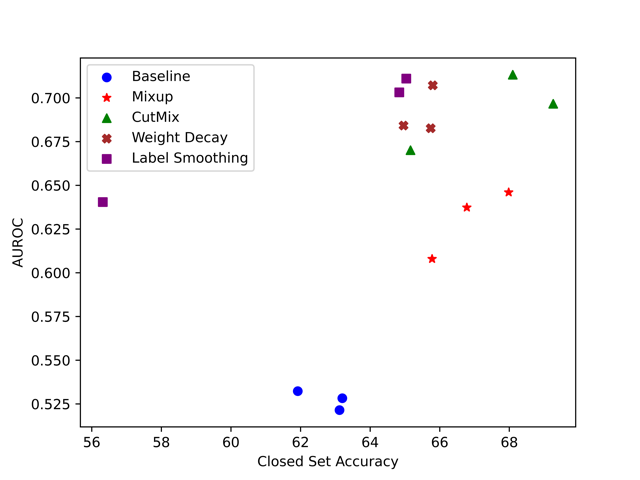}
    \caption{Closed-set accuracy vs AUROC for TIN~100-100 across three runs.}
    \label{fig:accVaurocTin}
\end{figure}

\cref{tab:auroc} shows the AUROC values for the different datasets and models.
Unlike the overlap values, a higher AUROC value corresponds to better OSP.
We again see that the use of these common regularization techniques helped OSP across the datasets.
Label smoothing on FGVCA~35-35 was the only regularizer that did not improve upon the baseline across all datasets and regularizers.
CutMix generally provided the greatest improvements over the baseline.
Notably, L2 regularization and label smoothing typically provided large boosts in OSP with relatively less gain in CSP compared to Mixup and CutMix.
Figures~\ref{fig:accVaurocCif} and \ref{fig:accVaurocTin} show that AUROC values for the baseline and regularization models are compact, having small relative standard deviations, from the three training runs.

\begin{figure}
\scriptsize
\begin{center}
\begin{tabular}{cc}
    \includegraphics[width=0.45\linewidth]{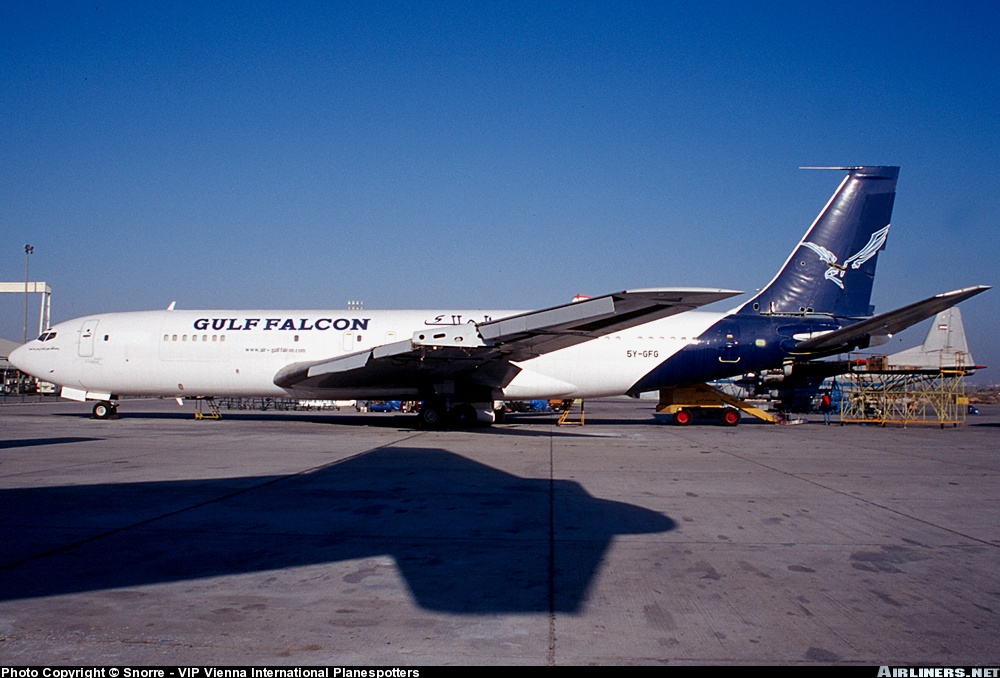} & \includegraphics[width=0.45\linewidth]{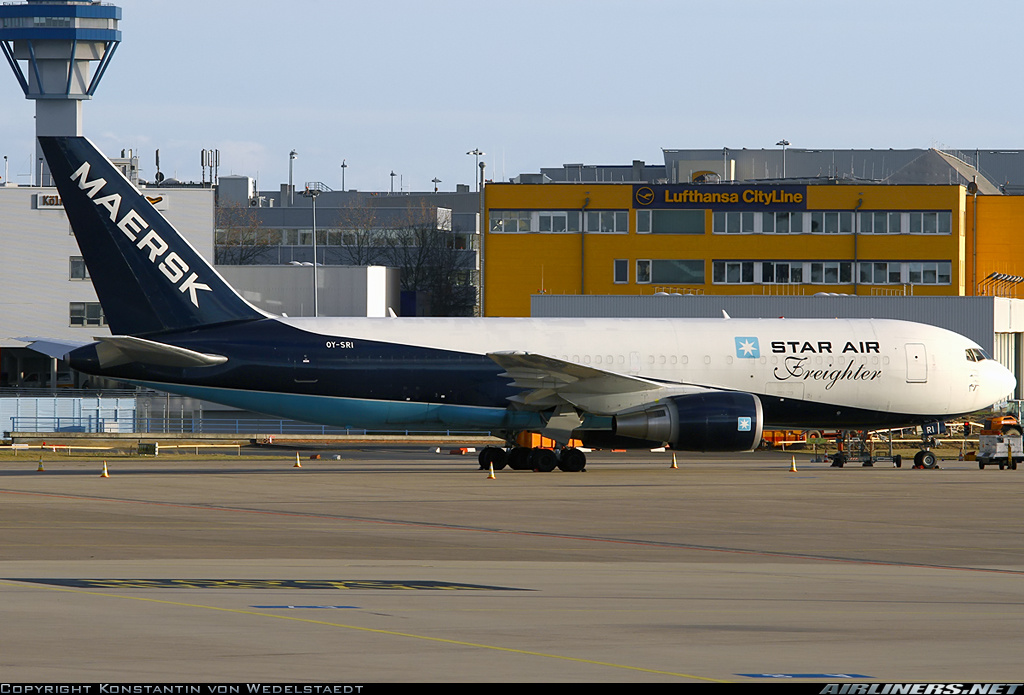} \\
    (a) FGVCA Boeing 707 aircraft. & (b) FGVCA Boeing 767 aircraft. \\ 
    \includegraphics[width=0.45\linewidth]{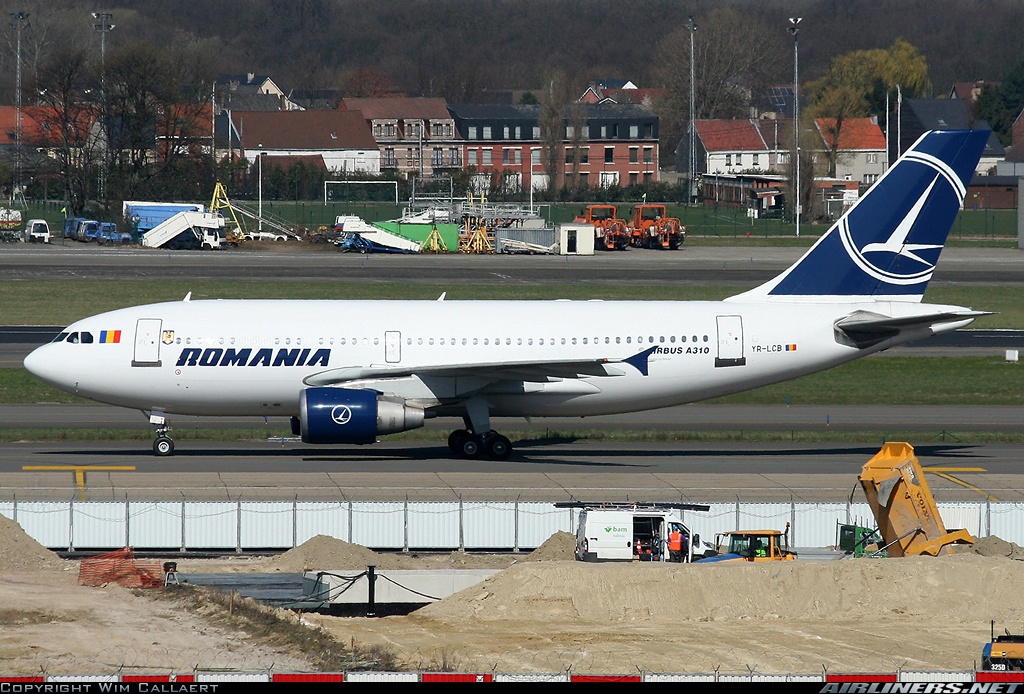} & 
    \includegraphics[width=0.45\linewidth]{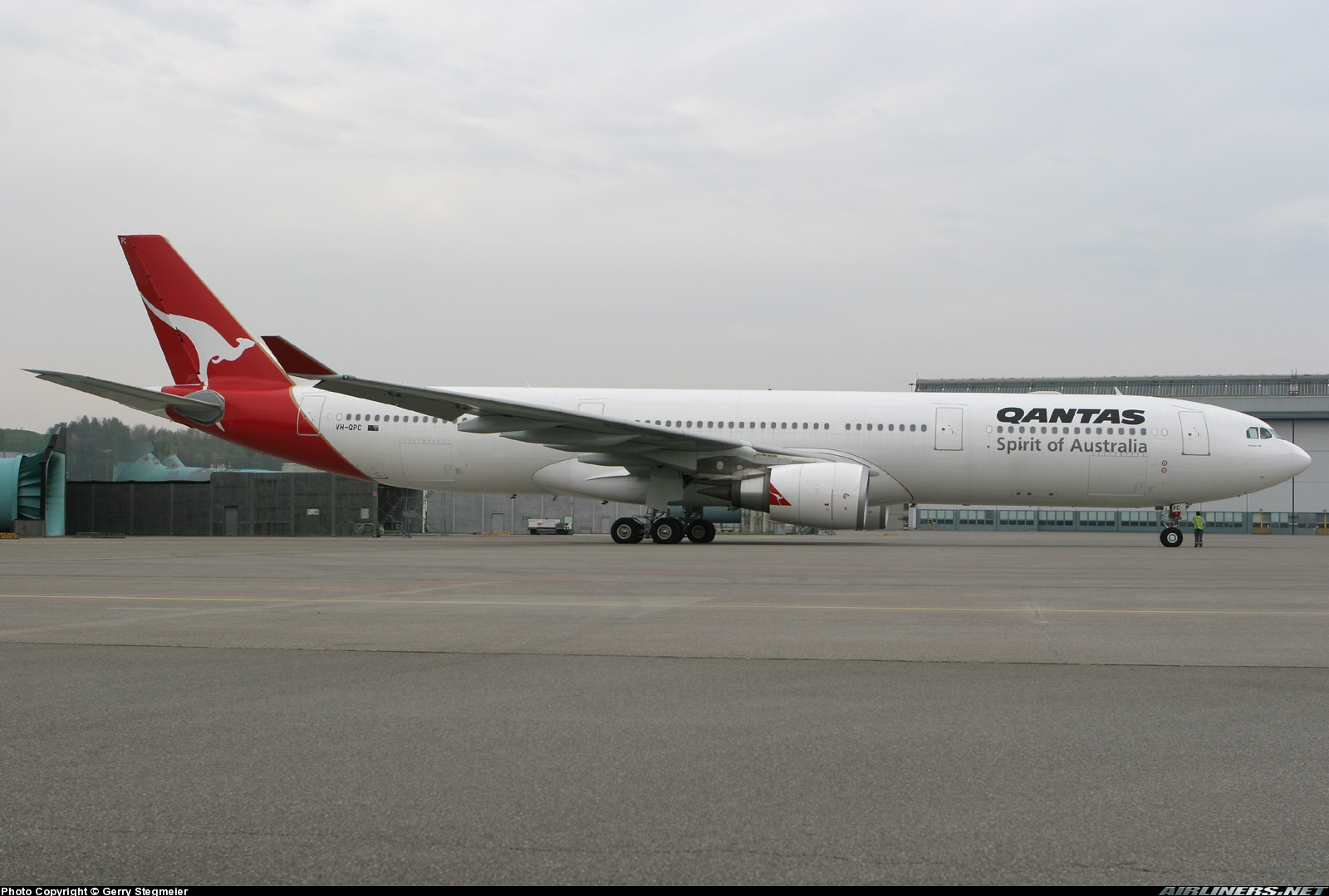} \\
    (c) FGVCA A310 aircraft. & (d) FGVCA A300 aircraft.
\end{tabular}
\end{center}
\caption{Example of visually similar FGVCA classes.} \label{fig:fgvca}
\end{figure}

The regularization schemes we tested provided improvements over the baseline model for both closed-set accuracy and OSP.
While there were slight improvements when using regularization with FGVCA~35-35, the improvements were minor in comparison to the other datasets tested.
We believe this was due to the relative difficulty of the open-set problem when using the FGVCA~35-35 dataset. 
Specifically, the center crop augmentation used only with FGVCA~35-35 could remove distinguishing features from the images. 
Additionally, as FGVCA 35-35 is a fine grained dataset, many of the classes are visually similar. 
Figure~\ref{fig:fgvca} highlights the strong similarities amongst classes in FGVCA, showcasing potential difficulties of performing OSR with this dataset.

% \begin{table}
%     \centering
%     \scriptsize 
%     \begin{tabular}{|c?c?c|c|c|c|}
%         \hline
%           Dataset &  Base &    L2 &    LS &    MU &    CM \\
%                 \hline\hline
% CI 6-4      &  0.30 &  0.20 &  0.22 &   0.21 &    0.18 \\ \hline
% CI 10+      &  0.36 &  0.15 &  0.20 &   0.31 &    0.32 \\ \hline
% CI 100+     &  0.27 &  0.14 &  0.15 &   0.16 &    0.12 \\ \hline
% TIN 100-100 &  0.35 &  0.18 &  0.22 &   0.35 &    0.29 \\ \hline
% FGVCA 35-35 &  0.24 &  0.24 &  0.23 &   0.22 &    0.21 \\ \hline
%     \end{tabular}
%     \caption{newbins overlap.}
%     \label{tab:overlap}
% \end{table}

\subsection{Stacked Regularizers}
\label{ssec:stacked}
In our experiments, we specifically employed one regularizer per training run to examine each technique's effects individually.
However, in practice multiple regularizers could be stacked to increase performance.
Hence, we also examined the effect of stacking regularization methods.
We performed one run each of stacked regularization schemes, one with CutMix and L2 regularization (CM + L2) and one with CutMix, L2 regularization, and label smoothing (CM + L2 + LS).
The results are shown in \cref{tab:stacked}. 
In this table, and all future tables where it applies, the arrow next to the metric indicates if a larger or smaller value for that metric is preferred.

As expected, the combination of regularizers outperformed the baseline model in closed-set accuracy and OSP.
Additionally, stacking regularizers outperformed using each regularization method independently on closed-set accuracy and OSP.

\begin{table}
    \centering
    % \scriptsize
    \resizebox{\columnwidth}{!}{
    \begin{tabular}{|c?c?c|c|c?c|c|}
    \hline
    Metric &   Base &     L2 &     LS &     CM &  CM + L2 &  CM + L2 + LS \\
                \hline\hline
  Accuracy $\uparrow$ &  92.97 &  94.76 &  93.16 &  95.27 &    96.67 &         96.47 \\
                \hline
 Overlap $\downarrow$ &   0.27 &   0.14 &   0.15 &   0.12 &     0.11 &          0.10 \\
                \hline
     AUROC $\uparrow$ &   0.65 &   0.87 &   0.80 &   0.89 &     0.89 &          0.90 \\
                \hline
    \end{tabular}
    }
    \caption{Performance values for one run of stacked regularizers on the CI~10+ dataset.}
    \label{tab:stacked}
\end{table}

\subsection{Adam Optimizer} 
\label{ssec:adam}
As described in \cref{sec:setup}, all models were trained using the SGD optimizer.
To test that our results hold not just for SGD, we revisited our approach using the Adam optimizer \cite{Adam}.
We performed one run of each regularizer and the baseline model on the CI~10+ dataset.
We also adjusted the learning rate from 0.1 to 0.001 as Adam typically favors smaller learning rates.
Results are shown in \cref{tab:adam}.
Predictably, using the regularization techniques with the Adam optimizer also helped to improve the CSP.
Despite using a different optimizer, we again found that the regularization techniques helped to improve the OSP.

\begin{table}
    \centering
    \scriptsize 
    \begin{tabular}{|c?c?c|c|c|c|}
                \hline
               Metric &   Base &     L2 &     LS &     MU &     CM \\
                \hline\hline
  Accuracy $\uparrow$ &  93.68 &  93.53 &  93.99 &  95.95 &  95.92 \\
                \hline
 Overlap $\downarrow$ &   0.22 &   0.15 &   0.14 &   0.15 &   0.11 \\
                \hline
     AUROC $\uparrow$ &   0.68 &   0.84 &   0.84 &   0.81 &   0.90 \\
                \hline
    \end{tabular}
    \caption{Performance values for one run of models trained with the Adam optimizer on the CI~10+ dataset.}
    \label{tab:adam}
\end{table}

\section{Analysis}
\label{sec:analysis}

In this section, we provide analysis on the relationship between CSP and OSP for regularized models. 
We also discuss how the feature spaces are changed as a result of regularization and how regularization relates to weight magnitude.
Throughout the following sections we italicize statements relating to the main contributions of the paper.

\subsection{Closed-Set vs Open-Set Performance Increase} 
\label{ssec:relationship}

In a previous work \cite{ClosedSet}, it was shown that the OSP of a model is strongly correlated with closed-set accuracy.
The authors found that there is a linear relationship between accuracy gains brought about by using larger models and OSP using basic maximum logit thresholding \cite{ClosedSet}.
In our work, we instead keep the model size fixed and examine the impacts of regularization on CSP and OSP.

One main difference between our findings and \cite{ClosedSet} is the nature of the relationship between closed and open-set performance.
They found that there is a linear relationship between gains in accuracy (with increasing model size) and gains in OSP.
\emph{However, we found that certain regularizers can have \underline{large} increases in OSP with \underline{little} increases in accuracy}.
This can be seen clearly in  \cref{fig:accVaurocCif}, where label smoothing has little effect on the overall accuracy but has significant impacts on the OSP.
This trend is also seen in the larger scale datasets like TIN~100-100 as shown in \cref{fig:accVaurocTin}.
We still observed that for a given accuracy there is a wide range of OSP values that is highly dependent on the regularizer employed.
For example, at an accuracy of roughly 66\%, there is a 0.1 gap in AUROC between the lowest-scoring regularizer (Mixup) and the highest-scoring regularizer (label smoothing).
\emph{In general, more accurate models do perform better in OSP.
However, the accuracy need not be correlated with increased OSP.}

As mentioned previously, the approach from \cite{ClosedSet} achieves higher accuracy by increasing the model size.
We next used a ResNet101 on the CIFAR10+ dataset to examine if the increase in parameters affects OSP.
The results for one run are shown in \cref{tab:larger}.
Despite the larger ResNet101 not improving in accuracy over the smaller ResNet18, the larger model was able to outperform the smaller model in OSP.
This not only substantiates the claim that accuracy need not be correlated with increased OSP, but additionally backs the theory that the number of parameters may have an impact on the OSP of a model.

\begin{table}
    \centering
    \scriptsize
    % \resizebox{\textwidth}{!}{
    \begin{tabular}{|c?c|c|}
    \hline
    Metric & Baseline ResNet18 &  Baseline ResNet101 \\
    \hline
    \hline
    Accuracy $\uparrow$ & 92.97 & 92.13 \\
    \hline
    Overlap $\downarrow$ & 0.27 & 0.23 \\
    \hline
    AUROC $\uparrow$ & 0.65 & 0.70 \\
    \hline
    \end{tabular}
    % }
    \caption{Performance values for one run of the ResNet101 model trained on the CI~10+ dataset compared to the average ResNet18 model.}
    \label{tab:larger}
\end{table}

\subsection{Feature Space Analysis}
\label{ssec:featurespace}

In the GAP feature space employed in this work, there will be some region of space that corresponds to each closed-set class. 
Three simple ways to increase the accuracy of the classifier would be to move these regions farther apart from each other, shrink the size of each region to create less overlap, or some combination of the two. 

\cref{tab:cos} shows the average cosine similarity between any pair of class prototypes of the closed-set.
This corresponds to measuring how far the classes move apart from each other after regularization.
\emph{In general, increased regularization decreased the similarity between class prototypes.}
This results in more space between class regions.

\begin{table}
    \centering
    \scriptsize 
    \begin{tabular}{|c?c?c|c|c|c|}
        \hline
            Dataset &  Base &    L2 &    LS &    MU &    CM \\
                \hline\hline
            CI 6-4      &  0.58 &  0.40 &  0.23 &   0.43 &    0.39 \\ \hline
            CI 10+      &  0.58 &  0.47 &  0.21 &   0.42 &    0.32 \\ \hline
            CI 100+     &  0.63 &  0.43 &  0.15 &   0.63 &    0.41 \\ \hline
            TIN 100-100 &  0.73 &  0.62 &  0.37 &   0.77 &    0.49 \\ \hline
            FGVCA 35-35 &  0.78 &  0.76 &  0.77 &   0.80 &    0.80 \\ \hline
    \end{tabular}
    \caption{Average cosine similarity amongst closed-set class prototype vectors where \underline{lower} is considered better.}
    \label{tab:cos}
\end{table}

\cref{tab:cosTarget} shows the average cosine similarity between closed-set examples and their target class prototype.
This corresponds to measuring how each class region ``shrinks''.
\emph{Here we observed that added regularization either kept the region sizes relatively consistent or decreased the region sizes compared to the baseline.}
Decreasing the class region sizes results in more empty space between the class regions.

\begin{table}
    \centering
    \scriptsize 
    \begin{tabular}{|c?c?c|c|c|c|}
        \hline
            Dataset &  Base &    L2 &    LS &    MU &    CM \\
                \hline\hline
            CI 6-4      &  0.87 &  0.95 &  0.92 &   0.85 &    0.87 \\ \hline
            CI 10+      &  0.84 &  0.95 &  0.92 &   0.84 &    0.85 \\ \hline
            CI 100+     &  0.70 &  0.82 &  0.71 &   0.70 &    0.67 \\ \hline
            TIN 100-100 &  0.77 &  0.85 &  0.71 &   0.75 &    0.72 \\ \hline
            FGVCA 35-35 &  0.92 &  0.91 &  0.86 &   0.92 &    0.90 \\ \hline
    \end{tabular}
    \caption{Average cosine similarity between closed-set examples and their target class prototypes where \underline{higher} is considered better.}
    \label{tab:cosTarget}
\end{table}

Combining the two previous results shows how adding regularization can help to increase OSP.
\emph{By moving class regions farther apart and potentially tightening the regions, we leave more space between the classes for open-set vectors to lie.
This in turn led to better separability of the open-set from the closed-set.} 
We confirmed this phenomenon by measuring the average cosine similarity between open-set examples and the class means. 
\emph{We observed in \cref{tab:cosOpenSet} that the open-set vectors moved farther away from the closed-set class regions as regularization was applied.}

The effect of separating and contracting class regions can be expected from applying label smoothing.
In a previous work \cite{whenLabelSmooth} it was shown that label smoothing causes training data to become much more tightly bound and for class means to become farther apart in the penultimate layer feature space. 
However, elements of label smoothing can also be found in the other regularizers examined in this work. 
In the formulation of Mixup and CutMix, the target vectors are changed to be a convex combination of two, one-hot label vectors. 
This can be seen as a form of label smoothing, where the smoothing distribution is a one-hot vector, as opposed to uniform \cite{OnMixup}.
Therefore, some of the feature space properties coming from label smoothing are also shown in Mixup and CutMix. 

\begin{table}
    \scriptsize 
    \centering
    \begin{tabular}{|c?c?c|c|c|c|}
        \hline
            Dataset     &  Base &    L2 &    LS &     MU &    CM.  \\ \hline\hline
            CI 6-4      &  0.60 &  0.54 &  0.36 &   0.45 &    0.44 \\ \hline
            CI 10+      &  0.56 &  0.59 &  0.27 &   0.41 &    0.36 \\ \hline
            CI 100+     &  0.49 &  0.48 &  0.22 &   0.50 &    0.38 \\ \hline
            TIN 100-100 &  0.61 &  0.66 &  0.40 &   0.64 &    0.49 \\ \hline
            FGVCA 35-35 &  0.77 &  0.75 &  0.75 &   0.80 &    0.79 \\ \hline
    \end{tabular}
    \caption{Average cosine similarity between open-set examples and all target class prototype vectors where \underline{lower} is considered better.}
    \label{tab:cosOpenSet}
\end{table}

\subsection{Weight Magnitude}

L2 regularization has the explicit goal of minimizing the weight magnitudes in a network (see \cref{eq:ssw}).
As L2 regularization helped to increase the CSP and OSP, it can be theorized that smaller weight magnitudes could be correlated with better OSP.
Therefore, we ask if the other regularizers examined in this work have a similar effect as L2 regularization on the network weights.
One way to measure the weight magnitudes in a network is to use the sum of squared weights (SSW), defined as 

\begin{equation}
    SSW = \sum_i w_i^2
    \label{eq:ssw}
\end{equation}

\noindent for all weight values in the network (excluding biases).

In \cref{tab:ssw} we show the SSW values for a ResNet18 trained with the various regularizers on the CI~10+ dataset.
We observed that there is no clear trend between SSW and open-set metrics.
CutMix and Mixup both \emph{increased} the SSW, while label smoothing and weight decay \emph{decreased} the SSW relative to the baseline model.
However, all 4 regularizers showed improvements in OSP over the baseline regardless of the increase or decrease of the SSW value.
This shows that while reducing the SSW can potentially help with improving OSP, it alone is not required.

\begin{table}
    \centering
    \scriptsize
    \begin{tabular}{|c|c|}
        \hline
        Method & SSW \\
        \hline
        \hline
        Baseline & \multicolumn{1}{|r|}{35,539} \\
        \hline
        L2 Regularization & \multicolumn{1}{|r|}{1,431} \\
        \hline
        Label Smoothing & \multicolumn{1}{|r|}{26,618} \\
        \hline
        Mixup & \multicolumn{1}{|r|}{39,057} \\
        \hline
        CutMix & \multicolumn{1}{|r|}{39,630} \\
        \hline
    \end{tabular}
    \caption{Average SSW values for ResNet18 models on CI~10+.}
    \label{tab:ssw}
\end{table}

%------------------------------------------------------------------------
\section{Conclusions and Future Work}
\label{sec:conclusions}

In this work, we explored how common regularization techniques affect the Open-Set Recognition task.
We introduced several methods for evaluating Open-Set Recognition effects of regularizers.
We empirically showed that the regularization techniques examined significantly improve the open-set performance of models. 
We also demonstrated that regularization causes class regions to shrink and move apart from each other. 
Finally, we showed that the gains in open-set performance due to regularization are not linearly correlated to closed-set accuracy.
Finally, we showed that reducing the SSW is not required to improve OSR.
In future work, we plan to use these regularization insights to design new loss functions and methods that would be favorable for Open-Set Recognition.

\section{Acknowledgements}

This work was supported in part by the U.S. Air Force
Research Laboratory under contract FA8650-21-C-1174.
Distribution A: Cleared for Public Release. 
Distribution Unlimited. PA Approval \#AFRL-2024-1693.
We also thank Logan Frank for comments on this work.

%%%%%%%%% REFERENCES
{\small
\bibliographystyle{ieee_fullname}
\bibliography{egbib}
}

\end{document}